\documentclass[letterpaper, 10 pt, conference]{ieeeconf} 
\usepackage{nameref}
\usepackage{hyperref}
\usepackage{graphicx}
\usepackage{doi}

\IEEEoverridecommandlockouts
\title{\LARGE \bf
A Novel Design and Evaluation of a Dactylus-Equipped Quadruped Robot for Mobile Manipulation 
}

\author{Yordan Tsvetkov$^{1}$ and Subramanian Ramamoorthy$^{2}$%
\thanks{$^{1}$ Yordan Tsvetkov is with the School of Engineering,
        University of Edinburgh, Old College, South Bridge, Edinburgh EH8 9YL
        {\tt\small s2026514@ed.ac.uk}}%
\thanks{$^{2}$ Subramanian Ramamoorthy is with the School of Informatics, University of Edinburgh,
        Old College, South Bridge, Edinburgh EH8 9YL
        {\tt\small s.ramamoorthy@ed.ac.uk}
}%
}

\graphicspath{{Images/}}
\begin{document}
\maketitle
\thispagestyle{empty}
\pagestyle{empty}
\begin{abstract}
Quadruped robots are usually equipped with additional arms for manipulation, negatively impacting price and weight. On the other hand, the requirements of legged locomotion mean that the legs of such robots often possess the needed torque and precision to perform manipulation. In this paper, we present a novel design for a small-scale quadruped robot equipped with two leg-mounted manipulators inspired by crustacean chelipeds and knuckle-walker forelimbs. By making use of the actuators already present in the legs, we can achieve manipulation using only 3 additional motors per limb. The design enables the use of small and inexpensive actuators relative to the leg motors, further reducing cost and weight. The moment of inertia impact on the leg is small thanks to an integrated cable/pulley system. As we show in a suite of tele-operation experiments, the robot is capable of performing single- and dual-limb manipulation, as well as transitioning between manipulation modes. The proposed design performs similarly to an additional arm while weighing and costing 5 times less per manipulator and enabling the completion of tasks requiring 2 manipulators.
\end{abstract}


\section{Introduction}
Legged robots are becoming increasingly more capable, and they are finding use in a variety of field applications requiring versatility and reliable mobility on rugged terrain. Quadruped (i.e. four-legged) robots have already been researched for use in a number of different applications \cite{bigdog,cheetah3,hyq}. 
Two of the more popular quadruped robot platforms that have become mature enough to be commercialised (Boston Dynamics Spot \cite{spotsite} and ANYbotics ANYmal \cite{anymalhard}) are marketed with inspection as a primary motivation. This means that their use is aimed at automating some of the tasks being performed by the human staff, requiring interactions with levers, handles, buttons and obstacles. This emerging need for mobile manipulation, even when the primary task is nominally only that of inspection suggests that there is a need for efficient integration of manipulation into the platforms. Currently, both of these platforms use optional 6- Degrees of Freedom (DoF) arms for manipulation tasks. The Spot Arm add-on is currently sold at the same price as Spot itself \footnote{Figure based on private correspondence.}.
\begin{figure}[t] 
\begin{center}
  \includegraphics[width=0.4\textwidth]{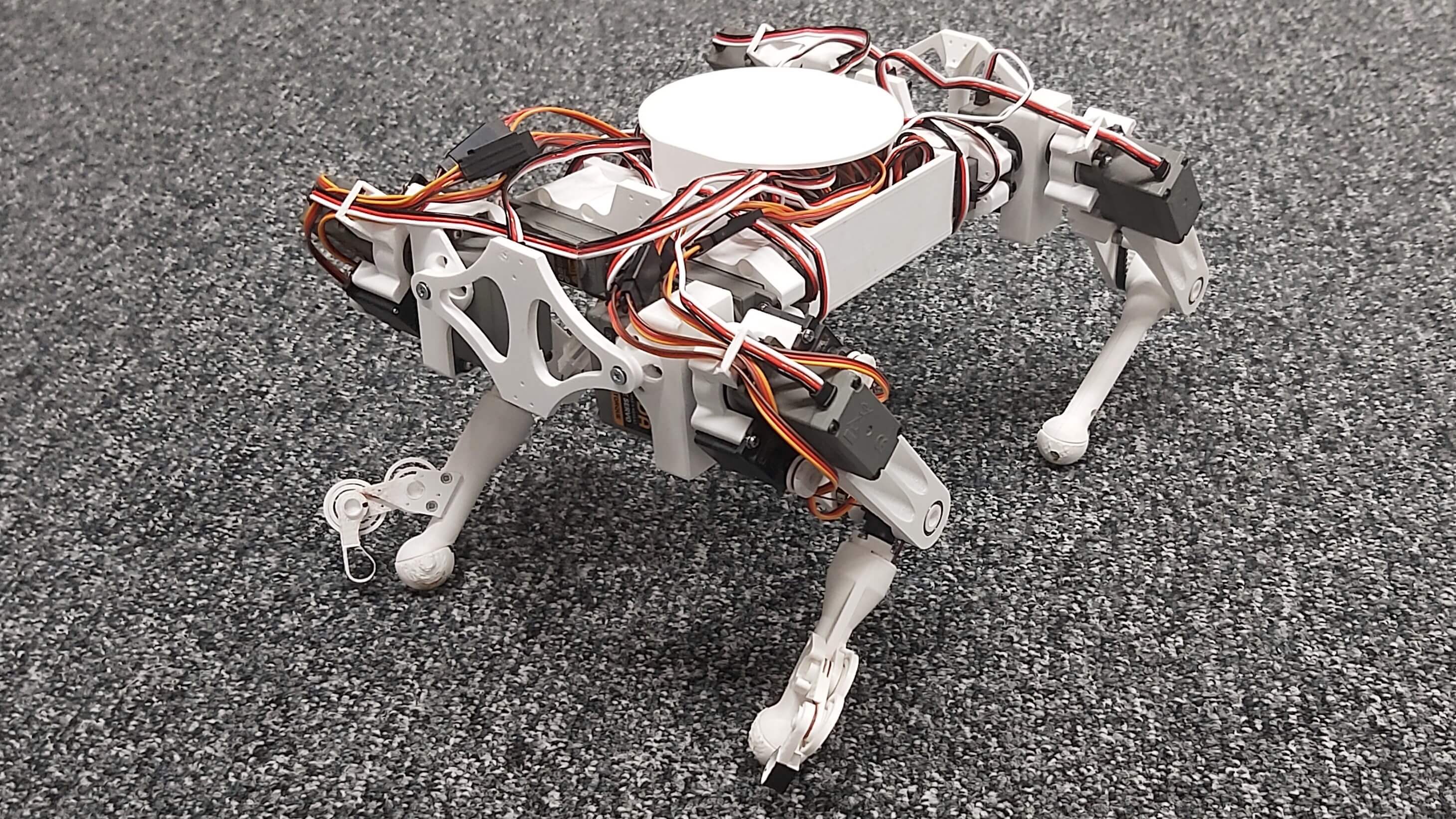}
  \caption{Our quadruped platform.}
  \end{center}
\end{figure} 
Within the walking robots community, much of the recent focus has been on achieving dynamic gaits and manoeuvres \cite{dynamicexamples2} required for successfully traversing unknown terrain, while simultaneously overcoming environmental perturbations \cite{robustwalking}. The physical requirements of these tasks have resulted in leg designs that can also apply sufficient precise force for dynamic and agile movement \cite{cheetah3,cheetahcub}.

The idea of using legs directly in the manipulation process has not been explored nearly as much in the literature \cite{anymalwithprongs,alphred}. On the other hand, {\textit{biological}} mobile manipulation makes significant use of tight integration between locomotion and manipulation.  Many arthropods make use of their front limbs to hold food for easier feeding, demonstrating such capabilities to an extent \cite{arthropodgrasping,objwithouthands}, while knuckle-walkers such as great apes and anteaters use their forelimbs for both walking and a variety of manipulation tasks such as grasping and foraging termite nests \cite{knucklewalking,anteater}. 

Arthropods have been used previously as an inspiration for the mechanical design of state-of-the-art robots \cite{insect1,insect2,insect3}. In our case, we are inspired by the biomechanics of the dactylus, a crucial feature of a crab's cheliped (leg adapted for grasping). Chelipeds are specialised for grasping \cite{arthropodgrasping} and possess a static section (propodus) and a movable part (dactylus). Based on this concept, after introducing an additional DoF in the dactylus and a wrist actuator for rotation axial to the tibia, a lightweight manipulator design is created that can successfully complete a variety of tasks. 

From an engineering perspective, we note that the legs in a quadruped are already designed for high carrying capacity \cite{legs_review}. There is also the possibility of using one or even two legs for manipulation without compromising stability, utilising the agility in the remaining body DoF to support this. So, it becomes feasible to explore the possibility of whole-body strategies for manipulating heavier objects without needing an additional limb. With the addition of smaller and less expensive actuators for the purpose of grasping, a robot's legs could potentially perform similar tasks as an arm attachment, representing useful reduction in cost. Equipping multiple legs with manipulators could also open entirely new possibilities for coordinated manipulation across multiple limbs \cite{billard2019trends}.

This paper reports on the design of a prototype small-scale quadruped robot with novel dual manipulators built into the legs. We then evaluate the physical capabilities of this robot to manipulate objects with its feet. This prototype is intended to explore design principles that could subsequently be transferred to larger-scale versions of this design, and also other platforms that currently do not make use of such integration between locomotion and manipulation.

In this paper, we make the following contributions:
\begin{itemize}
    \item A novel quadruped design is introduced, with manipulators built into the legs. This requires only 3 additional actuators per manipulator, representing a significant saving in contrast to externally mounted arms.
    \item Demonstration of control strategies for multiple types of manipulation that can be performed with such a robot (using one leg or two legs), including strategies for mode transitions, given high-level input from tele-operation (task level control and planning are deemed to be beyond the scope of this paper).
    \item Experimental evaluation of the hardware in a suite of tasks (also shown in the supplementary video clip), used to empirically demonstrate the capabilities of the robot.
\end{itemize}

The remainder of this paper is structured as follows. We first present the hardware design of the leg, followed by the specific design of the dactylus in \nameref{lwd} and the assembly in \nameref{ar}. Next, in \nameref{ex}, we describe the controllers needed for mobile manipulation. \nameref{RaD} presents experiments with a physical implementation of this design, followed by comparison to other platforms.

\section{Hardware Design}

A core contribution of this paper is the novel design of a quadruped robot, with the manipulation capability built into the legs directly. In this section, we describe the hardware design. Firstly, we describe the construction of a single leg of the quadruped. Next, we describe how this is modified to incorporate the actuators for manipulation. Finally, we describe how these are brought together in an integrated system. Our robot is a small quadruped research platform, with overall body dimensions 253 x 118 x 56 mm L x W x H. This is roughly comparable to a juvenile cat.

\subsection{Dactylus-Less Leg}
\subsubsection{Design}
\label{wo}
\begin{figure}[ht] 
\begin{center}
  \includegraphics[width=0.35\textwidth]{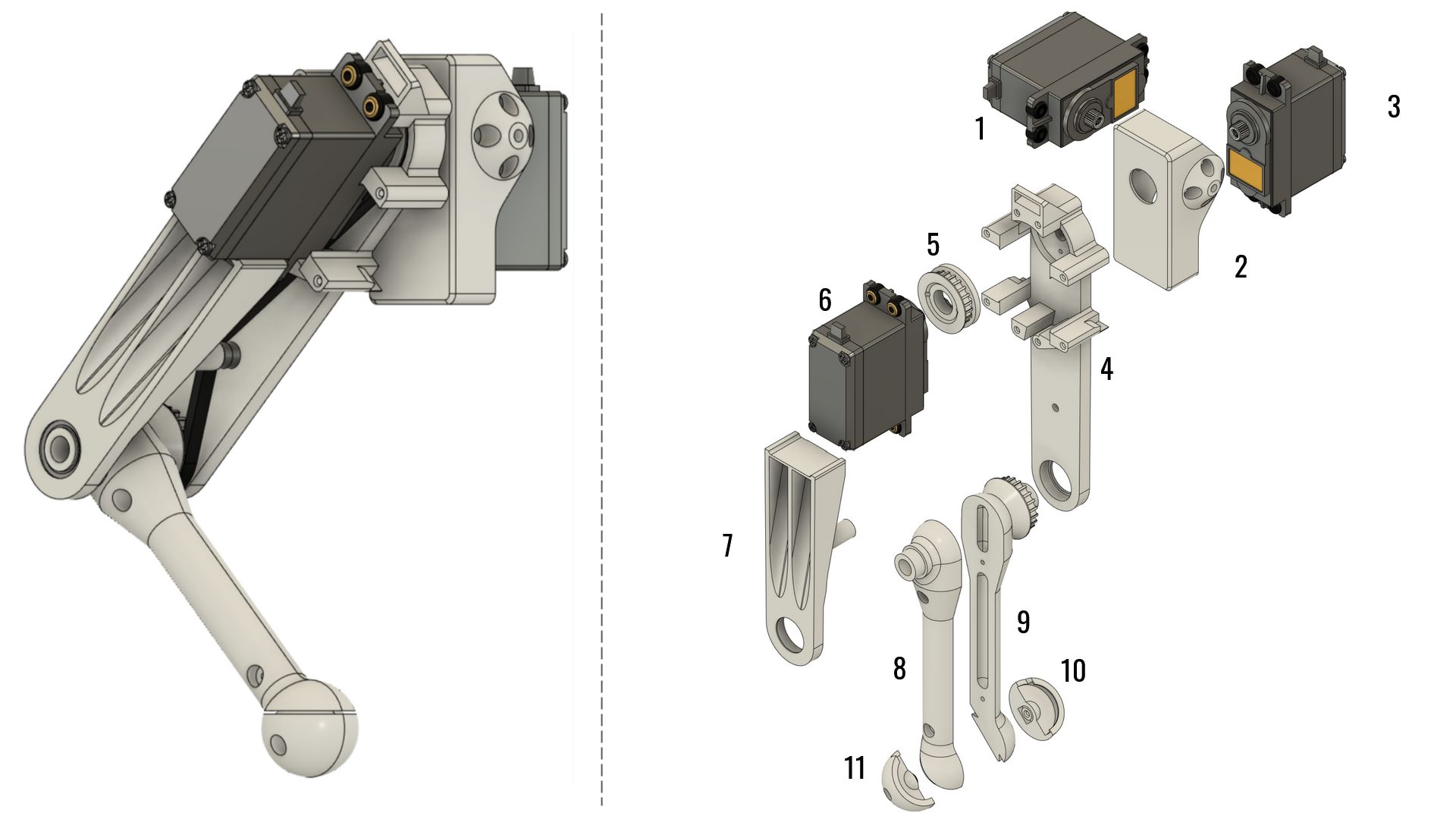}
  \caption{Leg without dactylus (left), exploded view of leg (right).}
  \end{center}
\end{figure}
The dactylus-less leg is designed with the goals of minimal weight and maximal agility. The femur (4) houses the tibia servo motor (6), which actuates the tibia (8-11) through a belt drive, tensioned with bearings mounted between the femur and femur support (7). The femur servo motor (3) is mounted on the coxa (2). The coxa servo (1) is static during actuation. This configuration allows the servo motors, which are the heaviest leg component, to be as close as possible to its axes of rotation, resulting in a significant reduction in moment of inertia (MoI).

The 3D-printed parts require no support structures, reducing printing time and material consumption. Assembling the leg has been streamlined compared to other quadrupeds thanks to the reduced component count. Other designs such as OpenQuadruped \cite{openquadruped} require additional parts to brace the servo motors and reduce the stress on their mounting points. Thanks to the attachment points for the tibia servo and the belt gear being close together, this is not a requirement for this design, further reducing its mass and complexity.\\
In order to further improve the performance of the leg, several optimisations have been made:
\begin{itemize}
    \item 4mm M3 belts are used instead of the more conventional 6mm M2 belts. This enables the servo pulley (5) to be mounted under the servo disk instead of over it, reducing design size and mounting point stresses.
    \item The tibia servo motor is integrated as a stress-bearing member, improving leg resistance to side forces.
    \item Cavities on the inner side of the tibia halves (8-9) improve their stiffness by causing additional load-bearing layers to be printed. Otherwise, these areas would be partially filled with infill, which is uniform and does not possess the directional stiffness of the cavities.
    \item Separating the femur into a thinner main femur (4) and a femur support (7) increases the inertial moment of the cross section by moving mass away from the axes of stress during walking without increasing the weight. Since the deflection $\theta$ of a beam is inversely proportional to its cross-sectional inertial moment $I_x$, this means that the leg will possess increased stiffness.
\end{itemize}
Thanks to these optimisations, the 3D-printed parts have a reduction of 53\% in total mass compared to OpenQuadruped when sliced (prepared for 3D-printing) at 15\% infill. Further comparison between the designs is done in \nameref{RaD}. 

\subsubsection{Actuation}
The leg uses 3 goBILDA 2000 Series servo motors for actuation. These servos were chosen for their high torque (approx. 25 kg-cm) and range of motion (300$^{\circ}$) and the leg has been designed according to their capabilities - both the femur and tibia servos make use of their full range, while the coxa has a range of 200$^{\circ}$ to prevent contact with the body. The chosen actuators are considered to have better range of movement and torque-to-price ratio in comparison to other motors of a similar scale. 

\subsection{Dactylus-Equipped Leg}
\label{lwd}
\subsubsection{Design}
\begin{figure}[ht]
\begin{center}
  \includegraphics[width=0.35\textwidth]{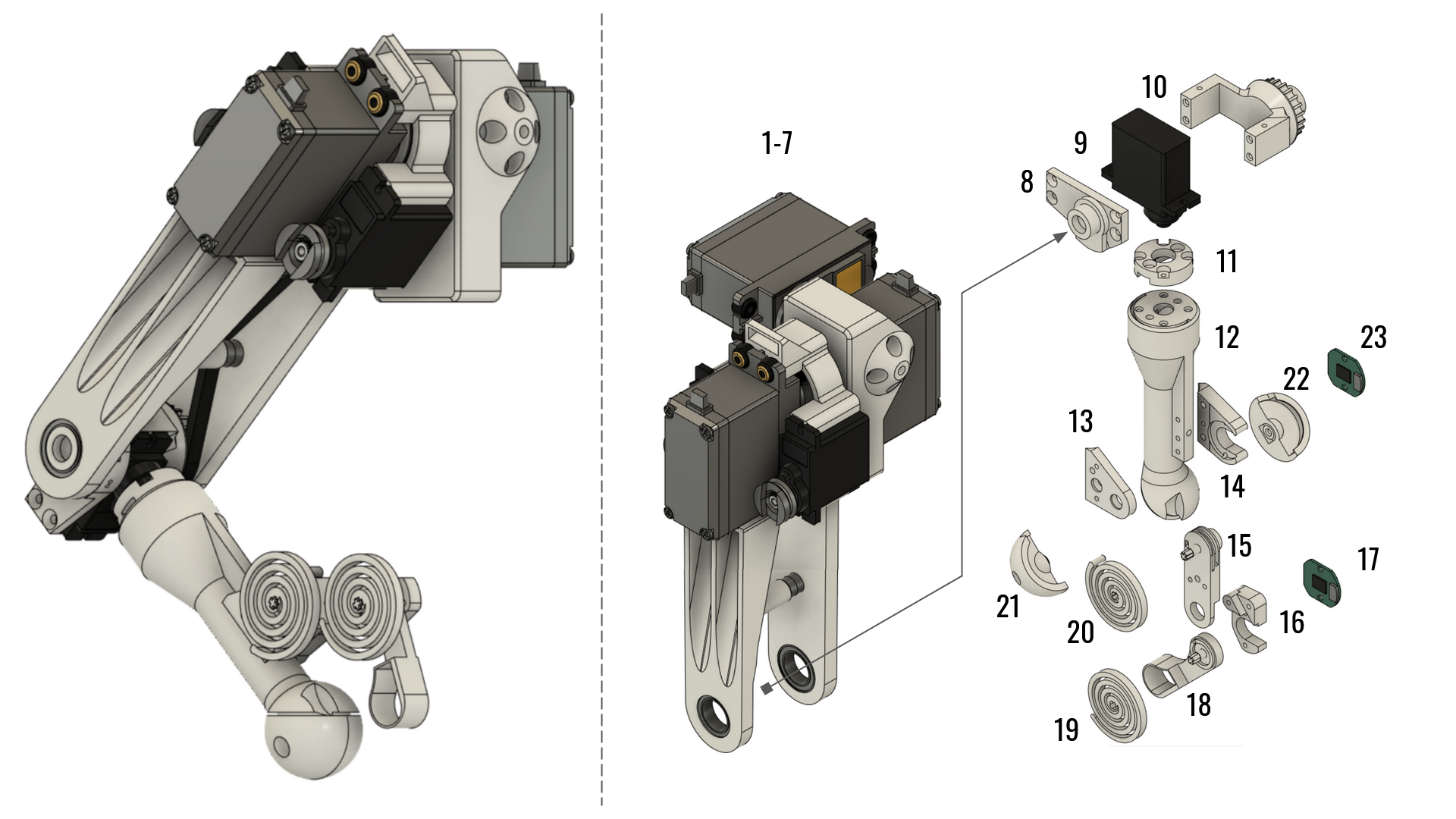}
  \caption{Leg with dactylus (left), exploded view of leg (right). Components 1-7 are same as leg without dactylus.}
  \label{fig:dactexploded}
  \end{center}
\end{figure}
The manipulation-enabled leg is inspired by a crab claw's anatomy with a dactylus and a static propodus (in this case, the tibia (8-12) serves the role of a propodus). It shares components 1-10 with the dactylus-less design. 

The dactylus itself (13-20) is designed to be assembled separate from the tibia so that replacing it does not require replacing the whole tibia. The tip of the dactylus is designed to be hollow, with the back of the tip being thicker to provide support during grabbing. This is inspired by the anatomy of the human fingertip, the thicker part of the tip having the same supporting role as a fingernail. 

To reduce MoI, the dactylus actuators are located on both sides of the tibia servo (as close to the axis of rotation as possible) and thanks to the weight savings detailed in \ref{wo}, the leg with all 3 additional actuators is 104.3 g lighter compared to a OpenQuadruped leg.

The physical properties of the legs are summarised in \nameref{comp}. Overall, the MoI of a fully extended leg has risen by approx. 58\%. This can be mainly attributed to the heavier tibia assembly (approx. 0.9E+05 g mm$^2$) and to its axial servo motor (approx. 1.47E+05 g mm$^2$). Despite this, the leg still possesses a smaller MoI about its servo motor axes compared to similar quadrupeds, as evaluated in \nameref{comp}.
\begin{figure}[ht]
\begin{center}
  \includegraphics[width=0.4\textwidth]{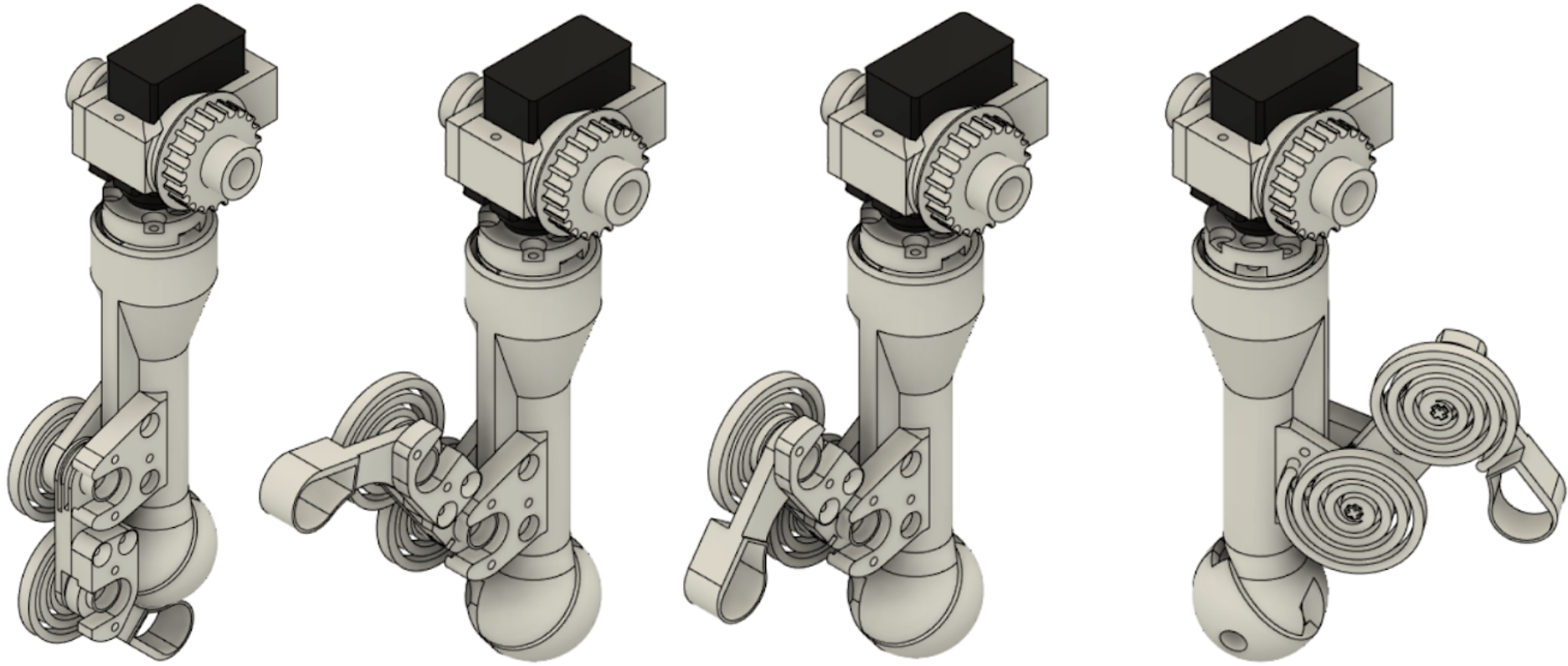}
  \caption{Degrees of freedom of dactylus.}
  \label{fig:dofs}
  \end{center}
\end{figure}

\subsubsection{Actuation}
There are 3 additional DFRobot Goteck 2.5kg micro servo motors that are responsible for manipulation - the base and tip dactylus motors (referring to which part of the dactylus they actuate; they are located on both sides of the tibia servo) and a motor (10) that enables rotation in the axis along the length of the tibia. 

The dactylus is actuated by spiral torsion springs and cables to enable compliance. The springs have the geometry of a logarithmic spiral, whose pitch, diameter and thickness are determined using a MATLAB nonlinear constrained optimisation script whose constraints are set by equations (9) and (10) in \cite{sts}. Although that original paper uses ABS springs, this paper adds to their results by demonstrating that the approach is also relevant for PLA springs. However, PLA is more prone to creep (deformation caused by constant stress below yielding point). As a result, the maximum stress of the springs have been set to be $\delta = 0.65$ times their yield stress instead of the value used in the paper ($\delta = 0.75$). 

\begin{figure}[ht]
\begin{center}
  \includegraphics[width=0.35\textwidth]{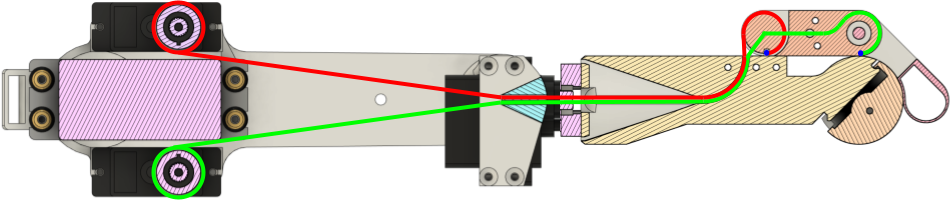}
  \caption{Cross section of leg showing cable trajectory.}
  \label{fig:crosssection}
  \end{center}
\end{figure}

The cables pass through a funnel-shaped channel in the tibia (see Fig. ~\ref{fig:crosssection}) to ensure that they retain their length when the tibia is actuated (hence, the dactylus retains its joint positions). Winches are used to connect the cables to the motors. They can have their compliance adjusted by changing their material or adding linear springs to them. Different kinds of fishing line can be used as tendons, since they are already manufactured to design requirements similar to ours. Braided fishing line is used in experiments not requiring force control, while monofilament fishing line has limited compliance and could allow for regulating the force applied on the manipulated object.

The tendons counteract spiral torsion springs in the dactylus joints (19-20 in Fig. ~\ref{fig:dactexploded}). By using rotary encoders to measure the dactylus joint angles, it is possible to calculate the reaction force $N_j$ for a joint $j$:
\[ N_j = k_t(r_1\alpha_1 - r_2\alpha_2) - k_s r_2\alpha_2, \]
where $k_t$ and $k_s$ are the spring coefficients of the tendon and spring, $r_1$ and $r_2$ the radii of the servo and dactylus winches and $\alpha_1$ and $\alpha_2$ the angular deviations measured by the servo and dactylus encoder respectively. By changing servo angles while holding an object, it is possible to regulate the force applied by the dactylus on this object.

\subsection{Assembled Robot}
\label{ar}
\subsubsection{Mechanical Design}
The body is composed of a torso that contains the electronics, 4 servo mounts and 2 brackets that counteract the torque on the coxa servo mounts. The body and brackets have been designed so that they can be printed at once on a standard 3D printer bed (220 x 220 mm), reducing manufacturing time. 
\subsubsection{System Architecture}
A Raspberry Pi 4 B+ performs the main calculations, while a Teensy 3.5 mounted on a PCB on top of the Raspberry Pi is responsible for low-level control and motor/sensor I/O. The PCB designed for the platform serves as a power distribution board and a breakout board, connecting the motors and sensors to the Teensy. The system can also function without the Raspberry Pi by receiving commands through the serial port of the microcontroller.

The robot can carry a BNO055 inertial measurement unit for orientation measurement, 4 Interlink Electronics 0.5 inch FSR (force-sensitive resistor) in the end effectors for measuring force and 4 magnetic encoders for dactylus position measurement when using compliant actuation cables.

\section{Trajectory Generation}
\label{ex}
Our novel robot design enables manipulation configurations involving the use of one or two of the legs while the rest of the body and legs are being used to stabilise and support this. In this section, we describe the low-level control strategies required to execute manipulation behaviours (based on tele-operated selection of targets).

\subsection{Jerk-limited Trajectory Generation}
Efficient manipulation is dependent on smooth yet high-speed motions. However, the desire for smoothness creates constraints on the dynamics of joint movements, which must be also set in accordance with the desire for speed. This calls for a dynamically-constrained joint trajectory generator.

\begin{figure}[ht]
\begin{center}
  \includegraphics[width=0.4\textwidth]{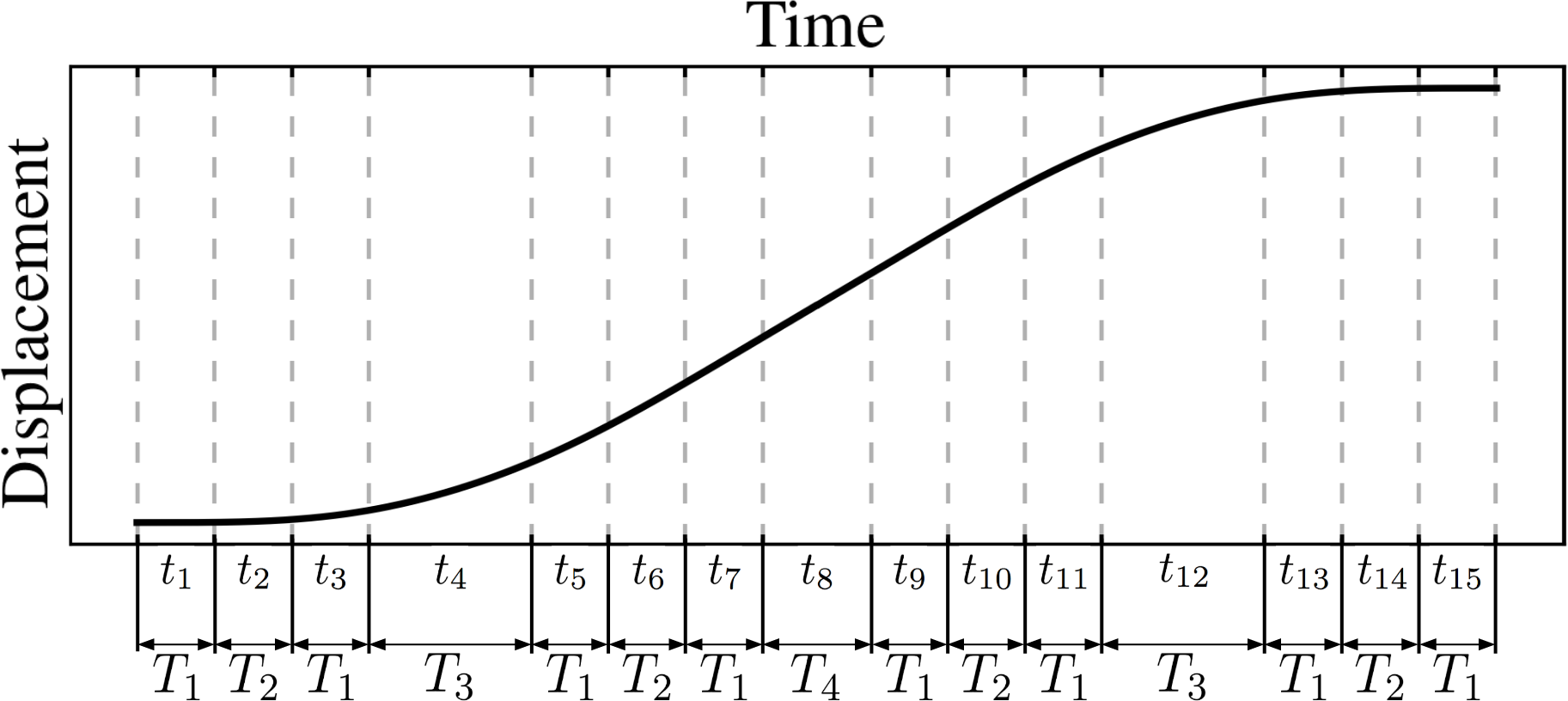}
  \caption{Example graph of trajectory displacement.}
  \label{fig:trajgraph}
  \end{center}
\end{figure}

For our experiments, we implemented a 15-phase jerk profile that produces a smoothened trapezoidal velocity profile \cite{15phase}. The phases $t_1-t_{15}$ are characterised by 4 durations $T_1-T_4$, as seen in Fig. ~\ref{fig:trajgraph}.
\subsection{Trajectory Generation for Single-leg Manipulation}
\label{singleleg}

Starting from an initial joint position $\phi_{0}$, the target position $\phi_{1}$ is received through serial communication with a notebook computer. Then, we check whether the acceleration and velocity limits $A_{max}, V_{max}$ will be reached in the motion. This is done by comparing the displacement $D = \phi_{1} - \phi_{0}$ and $V_{max}$ against the minimal displacements needed to reach $A_{max}$ and $V_{max}$ ($D_{aref}$ $D_{vref}$ resp.) and the min. velocity to reach $A_{max}$ ($V_{aref}$). One of 4 trajectory types is chosen depending on which limits are attainable, for which the respective durations $T_1-T_4$ are computed according to \cite{15phase}. 

After calculating $T_1-T_4$, a trajectory vector containing instantaneous position points is created by evaluating the displacement equations in \cite{15phase} for the phases $t_{1}-t_{8}$ (or to $t_{7}$ if $t_{8} = 0$) with a resolution of 1000 points/sec. Afterwards, the points are mirrored radially around the trajectory midpoint ($t_{8}/2$ or end of $t_{7}$ if $t_{8} = 0$). Finally, the trajectory is converted into PWM signals and sent to the motors. The equations for evaluating $T_1-T_4$ and the displacement can be found in \cite{15phase}.

\begin{figure}
\begin{center}
\vspace*{0.1in}
  \includegraphics[width=0.4\textwidth]{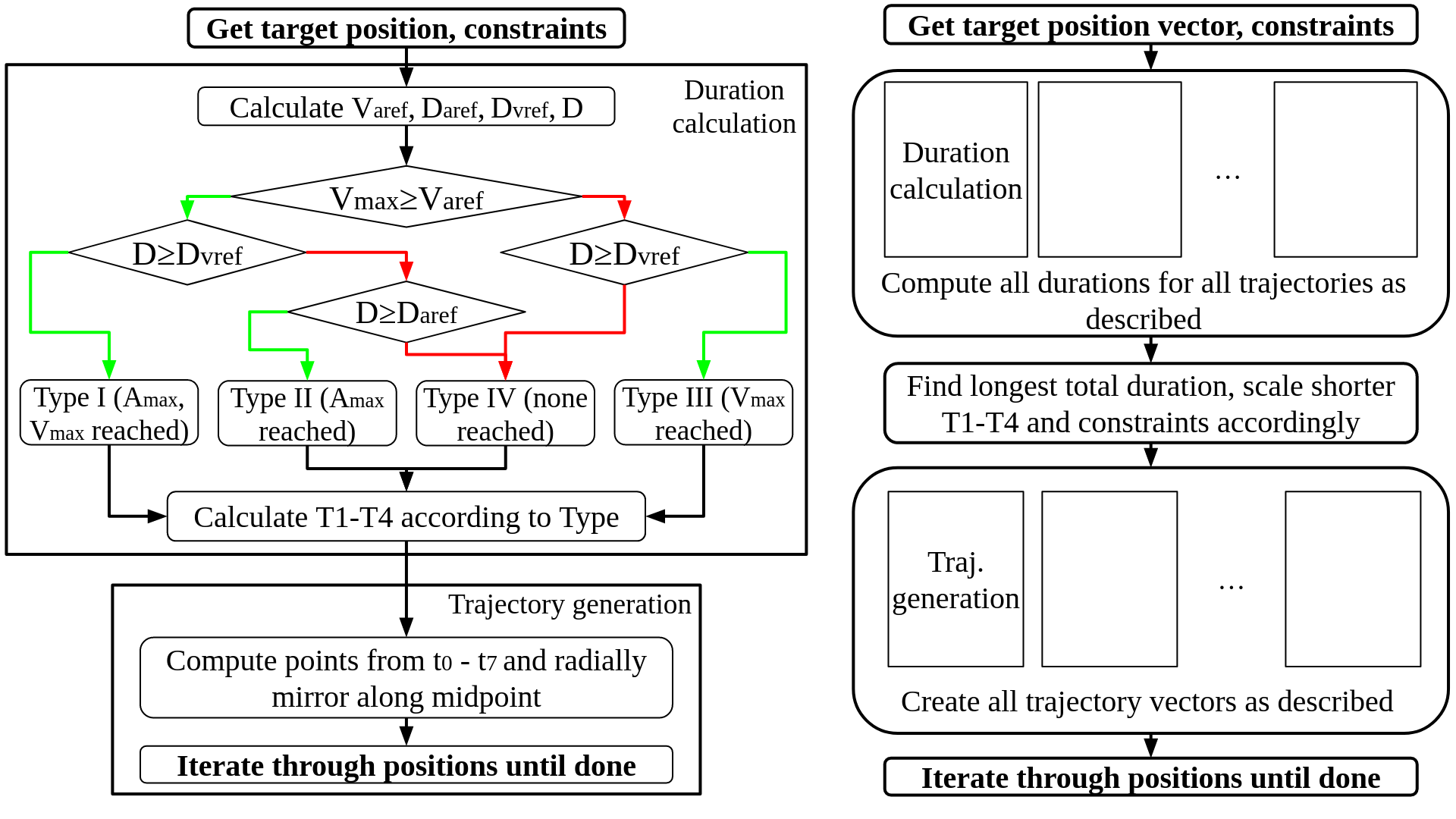}
  \caption{Block diagrams of single-manipulator and whole-body control.}
  \label{fig:diagrams}
  \end{center}
\end{figure}
\subsection{Synchronised Trajectory Generation}
\label{singletodouble}
Although the jerk-limited controller fulfils the requirements of speed and smoothness, in some cases coordinated, simultaneous motion of multiple DoFs is required. Hence, a synchronised trajectory controller is implemented.

After receiving a vector of target positions, $T_1-T_4$ is calculated for each $\phi_{1}$. The longest total duration $\sum_{i=1}^4 T_i$ is found and all other duration sets are scaled to match it. Additionally, the jerk constraint for each scaled movement must be divided by the respective scale factor in order to retain the correct final joint position \cite{15phase}. After these adjustments, trajectory vectors are generated for each $\phi_{1}$ and sent to the motors analogically to the single-leg controller.

\section{Experiments}
\label{RaD}
In the following experiments, we evaluate the physical capabilities of the robot, which is controlled via target position commands from a notebook computer. For trajectory generation, $J_{max} = 15$ rad/s$^3$, $A_{max} = 15$ rad/s$^2$ and $V_{max} = 5.2$ rad/s (the maximal servo motor velocity).
\subsection{Single Leg Manipulation}
\begin{figure}[ht]
\begin{center}
  \includegraphics[width=0.46\textwidth]{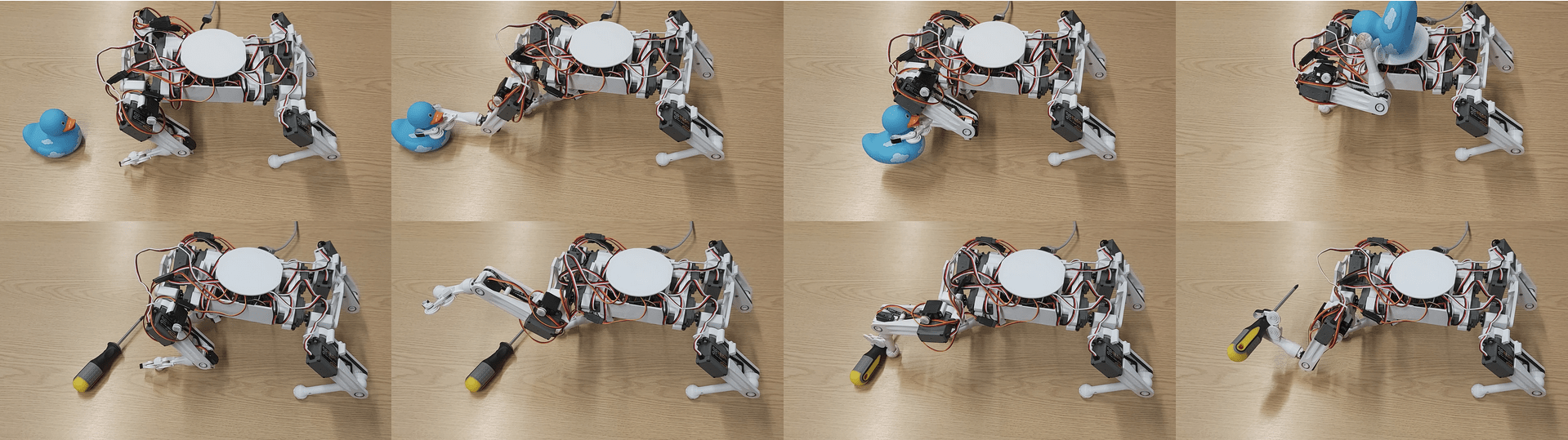}
  \caption{Manipulation of a rubber duck and a screwdriver.}
  \end{center}
\end{figure}
Using the jerk-limited controller presented in \nameref{singleleg}, the manipulation abilities of a single limb were evaluated. These experiments are aimed at demonstrating the full range of body capabilities of this novel mechanical design, using a range of manipulation tasks that would be of interest in field robotics. As is shown in the supplementary video, we have demonstrated the ability of this robot to perform object manipulation tasks including: (1) picking up and suitably orienting a 100 g screwdriver, (2) picking up, re-positioning and placing on the robot's backside, rubber ducks of different sizes. The screwdriver represents approx. 7\% of the robot's weight, while the ducks are used to demonstrate the range of sizes and complex shapes that can be handled by the manipulator. In the case of the ducks, single limb self-loading is possible. 

\begin{figure}[ht]
\begin{center}
  \includegraphics[width=0.4\textwidth]{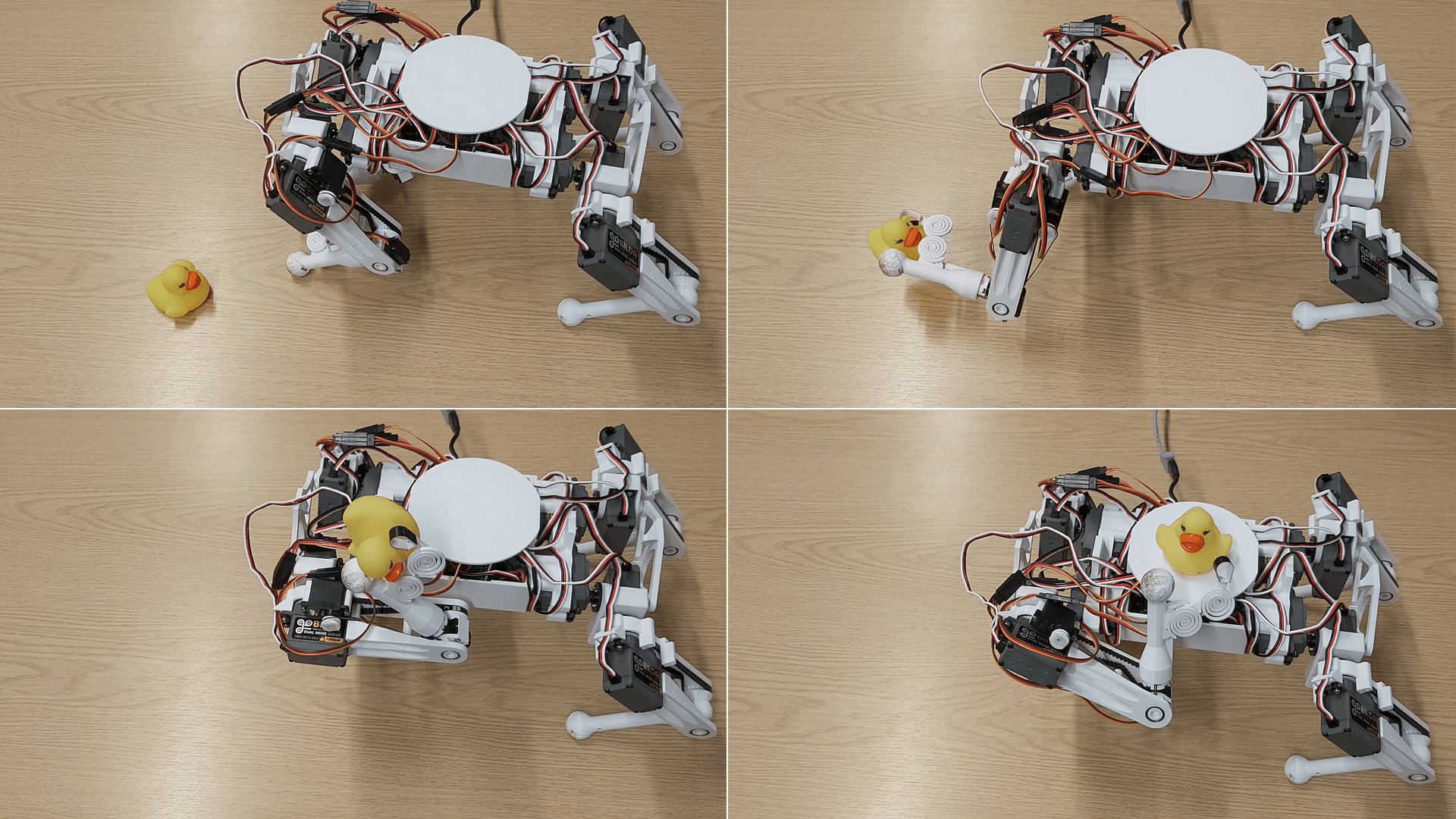}
  \caption{Self-loading of a rubber duck.}
  \end{center}
\end{figure}

\subsection{Transition Between Manipulation Modes}
Applying the controller presented in \nameref{singletodouble} allows for a smooth and stable transition between four-legged locomotion and static, dual-leg manipulation. Here, the robot computes its trajectory from 3 predetermined position vectors: (1) establishing a foothold on the tibiae of the hind legs, (2) rotation of the body upwards to position its centre of mass for vertical transition, and (3) body repositioning. As seen in the video clips, the robot is able to smoothly transition between the states without unnecessary joint stress while taking full advantage of motor capabilities. 

\subsection{Dual Leg Manipulation}
\begin{figure}[ht]
\begin{center}
  \includegraphics[width=0.4\textwidth]{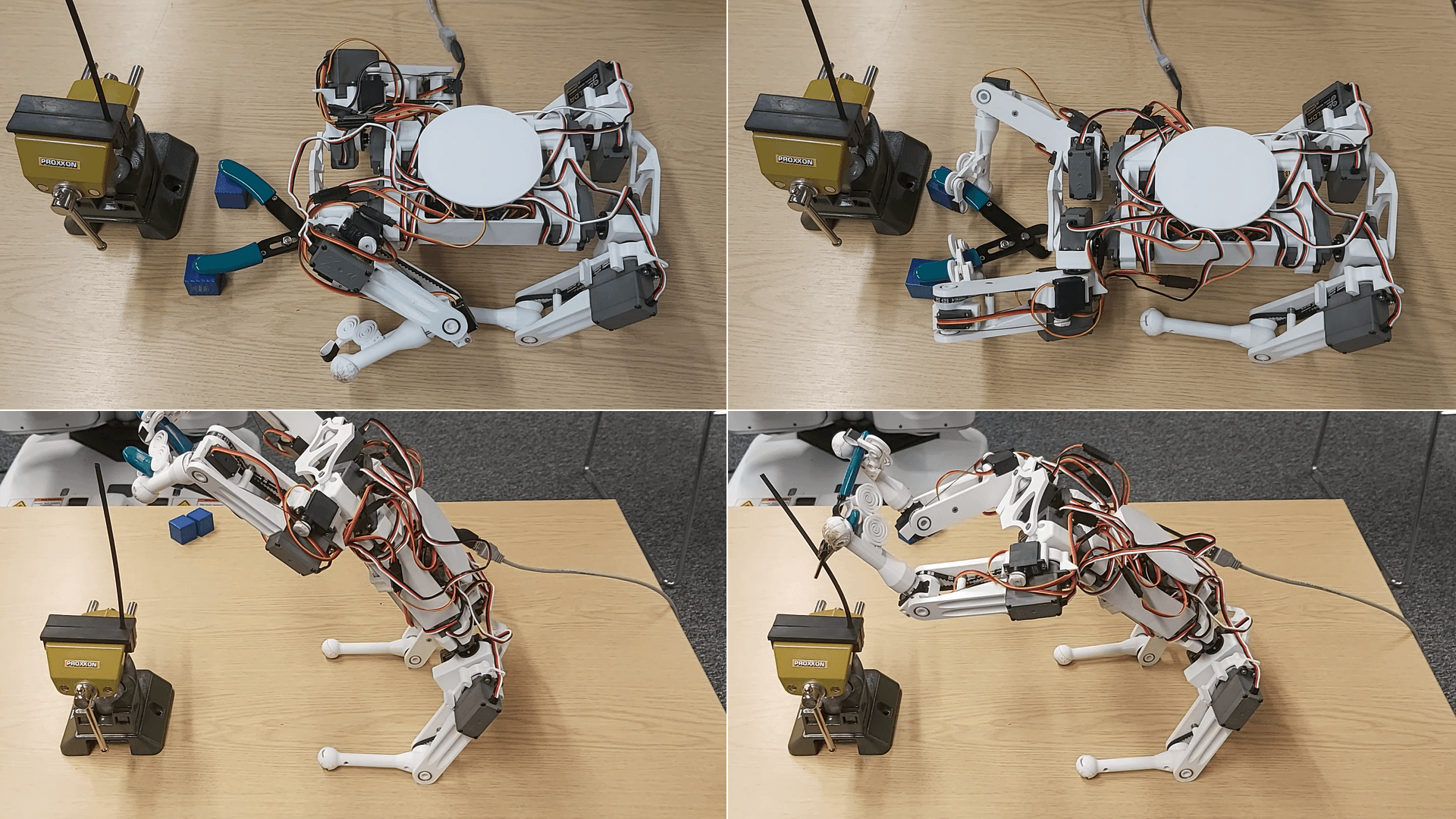}
  \caption{Stills of the dual-leg manipulation experiment.}
  \end{center}
\end{figure}
To explore dual-limb manipulation, we perform an experiment involving the usage of wire cutters to cut shrink tubing.

Using the tibiae for contact with the ground instead of the whole leg (which would be a more statically stable but less flexible configuration) enables actions that can extend the range of manipulation, namely whole-body movement. Since in this configuration the femur and body form a 2- DoF linkage, it is possible to perform movements that keep the centre of mass in a position over the support area while enabling the dactylus-equipped legs to reach new positions. 

The experiment requires whole-body motion for grabbing the cutters, and re-positioning to cut the tube. Also, there is a grasp strength requirement on the cutters for proper force application. As seen in the video, the robot has a sufficiently fine control of the cutters to position them for cutting. 

\section{Discussion}
\label{comp}
Among existing platforms, the one that is closest to our proposed design is OpenQuadruped \cite{openquadruped}, an open-source robotics research platform. Both use components with similar weights and have comparable leg lengths (200 vs. 248 mm). Even when normalised for the difference in length, the OpenQuadruped leg MoIs (9.54E+05 g mm$^2$ sagittal, 2.28E+06 coronal) are higher than those of a dactylus-equipped leg. Considering these factors, it can be observed that the negative effects of the inclusion of a dactylus have been mitigated via design optimisation. The comparison is summarised in the table below. The physical properties were evaluated from the CAD files of the designs. Here, `sagittal/coronal MoI' refers to the MoI about the femur and coxa motor axes respectively.

\begin{table}[h]
\begin{center}
\begin{tabular}{|c|c|c|c|}
\hline
Property & Our robot & OpenQuad & Unit\\
\hline
Femur Length & 100 & 106 & mm\\
\hline
Tibia Length & 100 & 142 & mm\\
\hline
Total Mass & 1481.6 & 2066.7 & g\\
\hline
Body Mass & 613 & 673 & g\\
\hline
Leg w/o Dact. Mass & 190.2 & 348.4 & g\\
\hline
Leg w/o Dact. Sagittal MoI & 5.39E+05 & 1.31E+06 & g mm$^2$\\
\hline
Leg w/o Dact. Coronal MoI & 7.99E+05 & 2.87E+06 & g mm$^2$\\
\hline
Leg w/ Dact. Mass & 244.1 & - & g\\
\hline
Leg w/ Dact. Sagittal MoI & 8.89E+05 & - & g mm$^2$\\
\hline
Leg w/ Dact. Coronal MoI & 1.21E+06 & - &  g mm$^2$\\
\hline
\end{tabular}
\end{center}
\end{table}

The table below shows an empirical comparison of a small-scale manipulator (Arm) and dactylus-equipped leg (Dact.) for our robot, as well as a larger-scale Kinova JACO2 6- DoF arm with a 1- DoF gripper, based on the configuration used in conjunction with an ANYmal robot in \cite{arm_anymal}, and an estimate for a dactylus-equipped ANYmal leg (ANYDact.). The estimates are based on the CAD files of our robot and the URDF models of ANYmal and JACO2.

\begin{table}
\begin{center}
\begin{tabular}{|c|c|c|c|c|c|}
\hline
Property & Arm & Dact. & JACO2 & ANYDact. & Unit\\
\hline
DoFs & 7 & 6 & 7 & 6 & N/A\\
\hline
Reach & 0.2 & 0.2 & 0.985 & 0.6 & m\\
\hline
Max. payload & 0.502 & 0.723 & 1.3 & 2.83 & kg\\
\hline
Mass added & 0.29 & 0.054 & 5.98 & 1.11 & kg\\
\hline
Total mass incr. & 19.5 & 3.77 & 20 & 3.8 & $\%$\\
\hline
Cost added & 105.5 & 20.28 & 26070 & 3692 & GBP\\
\hline
\end{tabular}
\end{center}
\end{table}

Since OpenQuadruped does not possess manipulation capabilities, a 6- DoF arm with a 1- DoF gripper has been designed. It has 3 goBILDA motors and 4 Goteck motors, the same models used in the dactylus-equipped legs. The presence of more motors further away from the base of the limb leads to a payload reduction.

The weight of a dactylus-equipped ANYmal leg is estimated from the dactylus/arm weight ratio of our robot (approx. 0.186) and the weight of the JACO2 arm. This leads to a dactylus weight equivalent to a 3- DoF KG-3 gripper. The payload estimates are conservative, as an unoptimised scenario of the whole mass being located at the tip of the leg is assumed. In a real-world dactylus, the motors would be close to the leg base similarly to our design.

In summary, dactyli possess a reduced weight and cost, making adding 2 dactyli feasible for performing complex tasks. The decrease in available DoFs can be offset by incorporating whole-body motions in manipulation as demonstrated in \nameref{RaD}. Although simultaneous locomotion and manipulation is not possible with dactyli, their dual-limb and whole-body manipulation abilities are more relevant to current research \cite{billard2019trends}. Furthermore, the potential mass reduction and freed up space at the top of the robot allow for bigger payloads to be transported, while allowing for self-loading and unloading to be conducted, as demonstrated.

\section{Conclusion}

This paper presented a novel, experimentally-validated approach to manipulation in quadruped robots by introducing a small-scale platform equipped with two biologically-inspired manipulators on its front legs. We showed that this robot is capable of reasonably complex tasks, utilising either one or two limbs to handle objects. To the best of our knowledge, this is the first quadruped platform capable of self-loading with a single leg and operating with handheld tools. This design is evaluated with reference to other platforms of roughly similar specifications, showing that dactylus-equipped limbs possess a MoI comparable to conventional quadruped limbs.

We view this as a first step in a programme of work around such dactyli. Our immediate next steps would be to further explore the range of dexterous manipulation. We are curious about the transfer of these advantages when the proof-of-concept design is scaled up, and we expect to be able to show that this can lead to even more efficient tendon routing. For example, using brushless motors with hollow shafts for the 'wrist' of the dactylus could allow the dactylus tendons to be centred and joint angle-agnostic without needing a complex tibia geometry, reducing weight and manufacturing costs. A larger platform could also demonstrate if dynamic locomotion would be affected by the inclusion of dactyli.

The added dexterity and strength of such a configuration could target manipulation-intensive applications in a wide variety of fields where quadrupeds can play a useful role. 
\bibliographystyle{IEEEtran}
\bibliography{bibliography}

\end{document}